\pdfoutput=1
\documentclass[11pt,a4paper]{article}
\usepackage[T1]{fontenc}
\usepackage[acceptedWithA]{tacl2021v1}
\usepackage{times,latexsym}
\usepackage{xurl}
\usepackage{booktabs}
\usepackage{multirow}
\usepackage{graphicx}
\usepackage{xcolor}

\newcommand{\AnnBfAdv}{4}
\newcommand{\AnnBfN}{6}
\newcommand{\AnnBfXs}{2}
\newcommand{\AnnKappa}{0.380}
\newcommand{\AnnKappaRef}{0.425}
\newcommand{\AnnN}{100}
\newcommand{\AnnNAdv}{90}
\newcommand{\AnnNXs}{10}
\newcommand{\AnnTargetedN}{20}

\newcommand{\EfourDetErrFour}{1.1}
\newcommand{\EfourHarmFour}{2.6}
\newcommand{\EfourHarmFourHi}{3.5}
\newcommand{\EfourHarmFourLo}{1.7}
\newcommand{\EfourHumRefFour}{3.6}
\newcommand{\EfourHumRefFourHi}{4.4}
\newcommand{\EfourHumRefFourLo}{2.8}
\newcommand{\EfourJudgeFour}{3.5}
\newcommand{\EfourKwFour}{4.7}
\newcommand{\EfourMinFisher}{0.14}

\newcommand{\EfourNoBfHarm}{2.6}
\newcommand{\EfourNoBfRef}{3.6}
\newcommand{\EfourNonHarmFour}{1.0}
\newcommand{\EfourPop}{12,480}

\newcommand{\EoneBCrossEightHi}{6.4}
\newcommand{\EoneBCrossEightLo}{0.8}
\newcommand{\EoneBCrossSixteenHi}{0.6}
\newcommand{\EoneBExceedN}{2}

\newcommand{\EoneBIdentEight}{0.0}
\newcommand{\EoneBModels}{6}
\newcommand{\EoneBRatioHi}{119}
\newcommand{\EoneBRatioLo}{57}

\newcommand{\EoneClozeMeanVsLetterRelN}{64}

\newcommand{\EoneClozeRobN}{1}
\newcommand{\EoneHaluAnyRob}{4}

\newcommand{\EoneHaluLetterOpp}{3}
\newcommand{\EoneHaluMeanSigFour}{6}
\newcommand{\EoneLetterRelVsLetterRevN}{80}
\newcommand{\EoneLetterRelVsLetterRevOpp}{4}

\newcommand{\EoneLetterRelVsLetterRevSignN}{43}
\newcommand{\EoneLlamaBBQCloze}{0.6}
\newcommand{\EoneLlamaBBQRel}{47.8}

\newcommand{\EoneMeanSigArcN}{2}
\newcommand{\EoneMeanSigN}{10}
\newcommand{\EoneOrderConsMean}{45}

\newcommand{\EonePosDownFour}{3}
\newcommand{\EonePosMaxFour}{11.0}
\newcommand{\EonePosSigEight}{3}
\newcommand{\EonePosSigFour}{8}
\newcommand{\EonePosTests}{16}

\newcommand{\EonePosUpFour}{5}
\newcommand{\EoneRepEightHi}{98.2}
\newcommand{\EoneRepEightLo}{89.6}
\newcommand{\EoneRepMin}{99.5}
\newcommand{\EoneRhoClozeLetter}{0.10}
\newcommand{\EoneRhoMeanSum}{0.60}
\newcommand{\EoneSigNClozeMean}{10}

\newcommand{\EoneSigNLetterRelNoTqa}{14}

\newcommand{\EoneSmolHaluCloze}{47.6}
\newcommand{\EoneSmolHaluRel}{97.5}
\newcommand{\EoneSmolHaluRev}{0.5}

\newcommand{\EoneTinyHaluCloze}{54.5}
\newcommand{\EoneTinyHaluRel}{18.8}
\newcommand{\EoneTinyHaluRev}{80.2}
\newcommand{\EoneTqaClozeSig}{0}
\newcommand{\EoneTqaClozeTests}{48}
\newcommand{\EoneTqaLetterSig}{10}
\newcommand{\EoneTqaLetterTests}{32}
\newcommand{\EthreeBoolqRemoved}{4}
\newcommand{\EthreeBoolqSig}{6}
\newcommand{\EthreeBoot}{5,000}
\newcommand{\EthreeChance}{4}
\newcommand{\EthreeEquivThree}{68}
\newcommand{\EthreeEquivTwo}{48}
\newcommand{\EthreeLlamaAUC}{0.77}
\newcommand{\EthreeLlamaDelta}{0.037}
\newcommand{\EthreeLlamaRecentred}{\ensuremath{-}0.008}
\newcommand{\EthreeLlamaYes}{0.48}
\newcommand{\EthreeLlamaYesq}{0.38}

\newcommand{\EthreeRepCells}{120}
\newcommand{\EthreeRepMean}{0.0035}
\newcommand{\EthreeSig}{11}
\newcommand{\EthreeSigBetter}{6}
\newcommand{\EthreeSigBinary}{9}
\newcommand{\EthreeSigWorse}{5}
\newcommand{\EthreeSmolAUC}{0.517}
\newcommand{\EthreeSmolAUCq}{0.553}
\newcommand{\EthreeSmolDelta}{0.154}
\newcommand{\EthreeSmolHi}{0.173}
\newcommand{\EthreeSmolLo}{0.130}
\newcommand{\EthreeSmolRecentred}{\ensuremath{-}0.031}
\newcommand{\EthreeSmolYes}{0.58}
\newcommand{\EthreeSmolYesq}{0.19}
\newcommand{\EthreeTests}{80}
\newcommand{\EthreeTrueYes}{0.62}
\newcommand{\GpBfAgreeMin}{96.9}
\newcommand{\GpBfCells}{14}

\newcommand{\GpBfSame}{12}

\newcommand{\GpCycDebRhoHi}{0.96}
\newcommand{\GpCycDebRhoLo}{0.68}
\newcommand{\GpCycDiff}{14}
\newcommand{\GpCycRho}{0.13}

\newcommand{\GpCycTests}{64}
\newcommand{\GpGenDeltaA}{9.2}
\newcommand{\GpGenDeltaB}{10.0}
\newcommand{\GpGenFlipEight}{35}
\newcommand{\GpGenFlipSixteen}{1}
\newcommand{\GpGenGeFlipEight}{50}
\newcommand{\GpGenGeFlipSixteen}{7}

\newcommand{\GpGenIdentEight}{4}
\newcommand{\GpGenIdentSixteen}{92}
\newcommand{\GpGenN}{520}
\newcommand{\GpGenOnlyA}{15}
\newcommand{\GpGenOnlyB}{20}
\newcommand{\GpGenQwIdentEight}{81}
\newcommand{\GpGenQwIdentSixteen}{99}
\newcommand{\GpGenRefGe}{90.8}
\newcommand{\GpGenRefQw}{99.8}

\newcommand{\GpLenChanged}{3}

\newcommand{\GpLenSmDropA}{18.8}

\newcommand{\GpPadEightHi}{5.6}
\newcommand{\GpPadEightLo}{2.2}

\newcommand{\GpPadN}{500}
\newcommand{\GpPadSixteenHi}{0.4}
\newcommand{\JudgeDisagreeMaxPP}{1.2}

\newcommand{\JudgeSignLit}{6}
\newcommand{\MechBatchAllHi}{0.6}
\newcommand{\MechBatchFourHi}{0.4}
\newcommand{\MechBatchN}{6}
\newcommand{\MechBatchSixteenHi}{0.6}

\newcommand{\MechBatchZeroHi}{5.8}
\newcommand{\MechBatchZeroPersistN}{5}
\newcommand{\MechPadAllHi}{0.4}
\newcommand{\MechPadFourHi}{0.2}
\newcommand{\MechPadN}{3}
\newcommand{\MechPadSixteenHi}{0.4}
\newcommand{\MechPadZeroHi}{7.4}

\newcommand{\MechSmolDefaultFlip}{4.8}

\newcommand{\MechSmolZeroFlip}{17.2}

\newcommand{\NModels}{8}
\newcommand{\PrecPub}{68.2\% (95\% CI 47.3--83.6)}
\newcommand{\PrecVtwo}{80.0\% (95\% CI 66.2--89.1)}
\newcommand{\PubVsVtwoPct}{91}
\newcommand{\RecallPub}{33.3\% (95\% CI 21.4--47.9)}
\newcommand{\RecallPubUntargeted}{60.0\% (95\% CI 40.7--76.6)}
\newcommand{\RecallVtwo}{80.0\% (95\% CI 66.2--89.1)}

\newcommand{\RvBothOppLet}{3}

\newcommand{\RvBothSameLet}{3}

\newcommand{\RvCeilHiLet}{0.83}

\newcommand{\RvCeilLet}{0.77}

\newcommand{\RvCeilLoLet}{0.69}

\newcommand{\RvCeilSum}{0.62}

\newcommand{\RvDiffHaluDeb}{6}

\newcommand{\RvDiffLet}{13}
\newcommand{\RvDiffSum}{6}
\newcommand{\RvDiffTests}{64}

\newcommand{\RvEitherAgreeLet}{10}

\newcommand{\RvEitherLet}{18}

\newcommand{\RvExDeb}{\ensuremath{-}7.3}
\newcommand{\RvExLet}{\ensuremath{-}13.6}
\newcommand{\RvExMean}{+9.8}
\newcommand{\RvExN}{2,000}
\newcommand{\RvExSigDeb}{significant}
\newcommand{\RvExSigLet}{significant}
\newcommand{\RvExSigMean}{significant}
\newcommand{\RvExSigSum}{not significant}
\newcommand{\RvExSum}{+0.8}

\newcommand{\RvKappaLet}{0.28}

\newcommand{\RvObsHarm}{4.9}
\newcommand{\RvObsHarmHi}{31.0}
\newcommand{\RvObsHarmLo}{\ensuremath{-}23.1}
\newcommand{\RvPosDown}{15}
\newcommand{\RvPosMaxData}{ARC-Easy}
\newcommand{\RvPosMaxFour}{59.5}
\newcommand{\RvPosMaxModel}{SmolLM2-360M}
\newcommand{\RvPosSig}{31}
\newcommand{\RvPosSigEight}{8}
\newcommand{\RvPosSigFour}{23}
\newcommand{\RvPosTests}{64}
\newcommand{\RvPosUp}{16}

\newcommand{\RvRelLet}{0.84}
\newcommand{\RvRelMean}{0.70}

\newcommand{\RvSplits}{1,000}
\newcommand{\RvTipHarm}{2.6}
\newcommand{\RvTipN}{51}
\newcommand{\RvTipRef}{3.6}
\newcommand{\SafeEightDeltaPP}{10.2}
\newcommand{\SafeEightModel}{SmolLM2-1.7B}

\newcommand{\SafeMaxDeltaPP}{18.5}
\newcommand{\SafeMaxFlipModel}{SmolLM2-1.7B}
\newcommand{\SafeMaxFlipPct}{28}

\newcommand{\SafeSigFour}{4}

\newcommand{\SigBBQFour}{2}

\newcommand{\SigHaluFour}{6}
\newcommand{\SigHaluFourUp}{3}

\newcommand{\WgCalHarmFour}{3.0}
\newcommand{\WgCalHarmFourHi}{3.7}
\newcommand{\WgCalHarmFourLo}{2.2}
\newcommand{\WgCalRefFour}{3.9}
\newcommand{\WgCalRefFourHi}{4.3}
\newcommand{\WgCalRefFourLo}{3.5}
\newcommand{\WgDHarmFour}{3.9}
\newcommand{\WgDRefFour}{4.2}
\newcommand{\WgNAnn}{86}
\newcommand{\WgPrecHarm}{78.6\% (95\% CI 60.5--89.8)}

\newcommand{\WgPrecLgHarmPt}{50.0}
\newcommand{\WgPrecRef}{83.3\% (95\% CI 70.4--91.3)}
\newcommand{\WgRecallHarm}{81.5\% (95\% CI 63.3--91.8)}

\newcommand{\WgRecallLgHarmPt}{55.6}
\newcommand{\WgRecallRef}{95.2\% (95\% CI 84.2--98.7)}
\newcommand{\WgRows}{23,280}
\newcommand{\WgRowsAdv}{12,480}
\newcommand{\WgRowsXs}{10,800}

\newcommand{\WgSigRefFour}{3}
\newcommand{\WgSignRefFour}{8}
\newcommand{\WgSmEightRef}{12.3}
\newcommand{\WgXsAnnAgree}{7}
\newcommand{\WgXsAnnN}{8}
\newcommand{\WgXsHi}{45.6}
\newcommand{\WgXsLlUp}{5.2}
\newcommand{\WgXsLo}{2.0}
\newcommand{\WgXsMaxEight}{3.6}
\newcommand{\WgXsPoolEight}{19.6}
\newcommand{\WgXsPoolFour}{19.1}
\newcommand{\WgXsPoolSixteen}{18.6}
\newcommand{\WgXsQwLgDown}{15.2}
\newcommand{\WgXsQwSmUp}{22.4}
\newcommand{\WgXsSafe}{250}
\newcommand{\WgXsSigFour}{4}
\newcommand{\WgXsSmDown}{7.2}

\title{Measurement Artifacts in Trustworthiness Evaluations\\
of Quantized Small Language Models}

\author{
  Tanzeel Abbas$^{1}$ \quad Tehseen Zia$^{1}$ \quad David Windridge$^{2}$ \\[2pt]
  $^{1}$COMSATS University Islamabad \quad $^{2}$Middlesex University London \\[2pt]
  \small\texttt{tanzeelabbas114@gmail.com} \quad \texttt{tehseen.zia@comsats.edu.pk} \\
  \small\texttt{d.windridge@mdx.ac.uk}
}

\date{}
\hypersetup{pdftitle={Measurement Artifacts in Trustworthiness Evaluations of Quantized Small Language Models},
  pdfauthor={Tanzeel Abbas, Tehseen Zia, David Windridge}}

\begin{document}
\maketitle

\begin{abstract}
Evaluations of quantized language models attribute the change in a benchmark score to
quantization. We write the measured change as the true change plus a term contributed by the
instrument. Instrument error that is the same at every precision can shrink a paired comparison but cannot create or reverse an effect. Error that interacts with quantization can do both, and it is reported as an effect of quantization. We estimate this interaction for four measurement choices in \NModels{}
instruction-tuned models below two billion parameters, at float16, LLM.int8() and NF4, on
identical items. Measured against labels from two human annotators, the 4-bit rise in harmful compliance is about half
to two-thirds of the rise in non-refusal that a corrected keyword detector records. Cloze and lettered
multiple-choice scoring disagree about the effect of quantization far more than their noise
allows: their estimates correlate at $\rho=\EoneRhoClozeLetter$, against a ceiling of about
\RvCeilLet{}. LLM.int8() answers and generations depend on which items share a batch, and
padding alone changes its answers. Rounding activations to 8 bits is enough to produce this dependence. NF4 is batch-invariant. The two largest reported
calibration losses are response-bias shifts. We release item-level predictions, option scores and
human annotations, and give recommendations for evaluating compressed small models.
\end{abstract}

\section{Introduction}

Small language models with fewer than two billion parameters are increasingly run on limited
hardware. Post-training quantization reduces the memory they need
\citep{dettmers2022int8,dettmers2023qlora}. A quantized model is usually accepted when its task
accuracy stays close to that of the original. Whether it remains well calibrated, truthful, safe
and fair is examined with standard benchmarks
\citep{hong2024decoding,xu2024beyondperplexity,lin2026slmtrust}.

Such evaluations compare a score before and after compression and attribute the difference to
quantization. We write the measured change as
$\Delta_{\mathrm{meas}} = \Delta_{\mathrm{true}} + \Delta_{\mathrm{inst}}$, where $\Delta_{\mathrm{inst}}$
is the change in the instrument's error between the two precisions. Instrument error that is the same at both precisions can shrink the measured change but cannot create or reverse it. Error that interacts with quantization can. A keyword detector that misses a phrasing of refusal which quantization makes more
frequent, or a scoring protocol that rewards a position preference which quantization
strengthens, adds a term that looks like an effect of quantization. Models below two billion
parameters are prone to such interactions. They often sit near floors and chance levels, phrase
refusals in ways that keyword lists do not anticipate, and are scored by detectors and prompting
protocols designed for larger models.

The interaction can decide the conclusion. On the \RvExN{} HaluEval items, the 4-bit change in the
accuracy of SmolLM2-1.7B is \RvExMean{} points under mean-normalised cloze scoring
(\RvExSigMean), \RvExSum{} under summed cloze scoring (\RvExSigSum), \RvExLet{} under lettered scoring in the released order (\RvExSigLet), and \RvExDeb{} when the lettered score is averaged over two option orders
(\RvExSigDeb), on the same model, items and precisions.

This paper estimates $\Delta_{\mathrm{inst}}$ for four choices that such studies make: how
refusals are detected, how multiple-choice items are scored, how inputs are batched, and how
calibration is summarised. We evaluate \NModels{} instruction-tuned models from five families,
between 135 million and 1.7 billion parameters. Each is evaluated at float16 and under the two
bitsandbytes configurations built into Hugging Face Transformers, LLM.int8() and 4-bit
NormalFloat (NF4). All comparisons are paired tests on identical items.

Our contributions are:
\begin{itemize}
\item A human-calibrated estimate of the effect of quantization on safety. After calibration
  against two annotators, 4-bit quantization raises harmful compliance by \EfourHarmFour{}
  points, or \WgCalHarmFour{} with WildGuard in place of Llama Guard. This is about half to
  two-thirds of the \EfourKwFour{}-point drop in refusal that a corrected keyword detector
  records. A tipping-point analysis shows how far this estimate depends on its main
  assumption.
\item A comparison of cloze and lettered scoring on the same items, models and precisions. NF4
  changes which answer position models prefer, and the two protocols disagree about the effect
  of quantization well beyond their measurement noise. The disagreement remains when the
  lettered score is averaged over all cyclic orders of the options.
\item Evidence that LLM.int8() is deterministic but not batch-invariant. In each of
  \EoneBModels{} models, the number of 8-bit answers changed by rearranging a batch is
  \EoneBRatioLo--\EoneBRatioHi\% of the number changed by quantization from float16. Padding
  alone also changes 8-bit answers, and 8-bit generations change with batch order in all three
  models tested. The dependence remains when the outlier decomposition is disabled and disappears when activations stay in float16, so rounding activations to 8 bits is enough to produce it. NF4 is batch-invariant.
\item A discrimination-aware analysis of calibration. The two largest reported calibration
  losses are response-bias shifts. Of \EthreeTests{} changes in ECE, \EthreeSig{} have
  uncorrected intervals that exclude zero, against about \EthreeChance{} expected by chance, and
  they have no consistent direction.
\item Recommendations for evaluating compressed small models, and a release of item-level
  predictions, raw option scores for every protocol, detector outputs and the adjudicated human
  annotations.
\end{itemize}
Table~\ref{tab:summary} in Section~\ref{sec:survives} summarises the main estimates.

\section{Related work}
\label{sec:related}

\paragraph{Trustworthiness under compression.}
Standard evaluations of compressed models can mislead. Perplexity misses capability losses that
knowledge-intensive tasks reveal \citep{jaiswal2024compressing}, and automatic metrics
underestimate quantization damage that human evaluators find \citep{marchisio2024multilingual}.
Broad benchmarks report small accuracy losses at 8 and 4 bits
\citep{li2024evaluating,kurtic2025bf16}. \citet{hong2024decoding} evaluate eight trust dimensions for compressed large models and find
moderate quantization largely benign. \citet{xu2024beyondperplexity} and
\citet{fu2025quantizeddeceptive} extend this to safety and to truthfulness under several
quantizers. Effects on social bias are mixed. \citet{goncalves2023compressionbias} find that
quantization can act as a regulariser that lowers bias, while later studies find effects that
differ by bias type \citep{marcuzzi2026quantization} and by demographic group
\citep{hua2026biasflipping}. \citet{paxton2026scalequantbias} show that apparent bias reductions
often coincide with more abstention. \citet{proskurina2024quantization} report that 4-bit GPTQ increases calibration
error and lowers confidence in correct answers, with effects that vary by model family and
scale. Safety has received most attention. \citet{chen2025qresafe}, \citet{kharinaev2025safety}
and \citet{kadadekar2026quality} find safety losses that depend on the model and, in part, on the
quantizer.
\citet{prasad2026quanttemp}, in contrast, find standard 4-bit quantization close to
safety-neutral for most models and larger effects from the sampling temperature.
On clinical benchmarks, 8-bit GPTQ preserves safety while 4-bit effects depend on the model and
the task \citep{twagirayezu2026clinical}. \citet{kurtskhalia2026determinism} finds that 4-bit
quantization of an 8B model concentrates its recommendations without shifting stereotype
direction. \citet{egashira2024exploiting} show that quantization can be exploited adversarially. Few of
these studies focus on small models. \citet{paxton2026scalequantbias} and \citet{lin2026slmtrust}
do, but neither audits the measurement instruments. We hold the
models and items fixed and vary the instrument.

\paragraph{Instance-level effects of compression.}
Aggregate accuracy hides which decisions compression changes \citep{hooker2019forget}.
\citet{dutta2024accuracy} show that quantized models flip many answers while accuracy holds, and
\citet{wu2026decisions} model how quantization shrinks the full-precision decision margin and
predict flip rates from it. \citet{felicio2026divergence} shows that, across compressed copies of
many models, the flip rate tracks the total variation distance between the output
distributions. We use paired
per-item tests throughout and ask whether the flips a study counts depend on its scoring
protocol.

\paragraph{Measurement validity of automated judges.}
A benchmark score measures an unobservable construct, and its validity has to be checked rather
than assumed \citep{jacobs2021measurement,wallach2025position}. When an automated instrument
replaces human labels, its misclassification biases the estimated rate. Epidemiology corrects
prevalence estimates for a test's known sensitivity and specificity
\citep{rogan1978estimating}, and prediction-powered inference corrects automated labels with a
small human-labelled sample \citep{angelopoulos2023ppi,boyeau2024autoeval}. We ask a narrower question of such
measurements: whether their error changes with precision. Keyword refusal detectors and model-based safety classifiers such as Llama Guard
\citep{inan2023llamaguard} disagree substantially with human labels
\citep{han2024wildguard,mazeika2024harmbench}, and string matching overstates the success of
jailbreaks \citep{souly2024strongreject}. The reliability of automated judges varies across
attacks and models \citep{schwinn2026coinflip}, and safety judges are sensitive to output style
and can themselves be attacked \citep{eiras2025knowthyjudge}. Jailbreak evaluators also
encode different definitions of success, so the same responses can yield very different attack
success rates \citep{gao2026jailbreakjudge,mu2026jailbreakevaluators}. SORRY-Bench compares
judge designs for refusal evaluation against human annotations \citep{xie2025sorrybench}.
\citet{nutfullin2026keywordscored} show that keyword scoring of multimodal models becomes invalid
when a defense writes safety language into the answers it evaluates. Closest to our analysis,
\citet{chen2026keywordsdrop} show that under KV-cache compression of Qwen2.5-3B,
keyword-detected refusal falls while classifier-judged refusal does not. We measure detector
error against human labels for weight-quantized small models and propagate it into the
estimated effect of quantization.

\paragraph{Option-order sensitivity.}
\citet{robinson2023leveraging} compare cloze scoring with prompts that list lettered options,
and \citet{gu2025olmes} find that small models often cannot use the lettered format and
recommend cloze scoring for them. Language models prefer particular option letters or positions
in multiple-choice formats \citep{zheng2024selectors,pezeshkpour2024order}, and small changes to a
prompt's formatting can change accuracy by many points \citep{sclar2024quantifying}. First-token letter
probabilities can disagree with the text a model would generate
\citep{wang2024myanswerisc,wang2024lookatthetext}, and \citet{ok2026onebiasscore} decompose
option bias into a flip rate and a positional skew. Small changes in choice order or answer
selection reorder leaderboards \citep{alzahrani2024benchmarks}, results depend on the evaluation
setup \citep{biderman2024lessons}, and casting choice tasks as completion reduces variance for
smaller models \citep{madaan2024variance}. Accuracy and order sensitivity do not move
together under debiasing \citep{hanna2026abcstable}, and the prediction of an answer letter is
localised in a few middle layers and attention heads \citep{wiegreffe2025answer}. These studies concern full-precision
models. Section~\ref{sec:protocol} asks whether quantization
changes such preferences and, through them, measured accuracy.

\paragraph{Calibration measurement.}
Expected calibration error (ECE) bins predictions by confidence and averages the gap between
confidence and accuracy \citep{guo2017calibration}. Its value depends on the binning scheme, and
\citet{nixon2019measuring} show that different calibration metrics can rank models differently.
ECE also combines two properties that compression may affect separately: how well confidence
separates correct from incorrect answers, and how far the predictions lean towards one label.
We report discrimination next to ECE and test whether a shift in label bias explains a change in
ECE.

\paragraph{Nondeterminism in inference.}
Floating-point addition is not associative, so the output of a forward pass can depend on how
kernels split their reductions, which in turn depends on batch size
\citep{he2025nondeterminism}. Changes in batch size, numerical precision and hardware can alter
generated text \citep{yuan2025nondeterminism}. These studies concern unquantized inference.
Section~\ref{sec:int8} shows that LLM.int8() adds a batch dependence of its own.

\section{Study design}
\label{sec:design}

\paragraph{Models and precisions.}
The \NModels{} instruction-tuned models are SmolLM2 at 135M, 360M and 1.7B
\citep{allal2025smollm2}, Qwen2.5 at 0.5B and 1.5B \citep{qwen2025}, Llama-3.2-1B
\citep{llama32,meta2024llama32}, Gemma-3-1B \citep{gemma2025} and TinyLlama-1.1B \citep{zhang2024tinyllama}.
Each is evaluated at float16, with LLM.int8() \citep{dettmers2022int8}, and with NF4 at a
float16 compute dtype \citep{dettmers2023qlora}, all through bitsandbytes. We write 8-bit for LLM.int8() and 4-bit for NF4. These are the
quantized loading options built into Hugging Face Transformers, and hence the default in
evaluation pipelines built on it. Our claims are restricted to these two configurations.

\paragraph{Baseline evaluation.}
Throughout the paper, the \emph{baseline evaluation} is our own first evaluation of all
\NModels{} models at all three precisions on every benchmark below, run once on a single cloud
GPU (NVIDIA A10G). It makes choices that are common in this literature: refusal is detected by a
keyword list applied to the first 200 characters of each response, multiple-choice items are
scored by mean-normalised cloze log-probability in fixed batches, and calibration is summarised
by ECE over ten bins. It is not a previously published study. Sections~\ref{sec:refusal}
to~\ref{sec:calibration} hold its models, items and outputs fixed and vary one of these choices
at a time.

\paragraph{A measurement model.}
For a binary outcome such as refusal, let $\pi_p$ be the true rate at precision $p$, and let the
instrument have sensitivity $s_p$ and specificity $c_p$ at that precision. The measured rate is
$m_p = s_p\pi_p + (1-c_p)(1-\pi_p)$. For float16 ($f$) and a quantized precision ($q$),
\[
\begin{array}{r@{\;}l}
m_q - m_f = & (s_f + c_f - 1)(\pi_q - \pi_f) \\
 & +\, (s_q - s_f)\,\pi_q \\
 & -\, (c_q - c_f)(1 - \pi_q).
\end{array}
\]
The first term rescales the true change by the instrument's discrimination. For an instrument better than chance ($s_f + c_f > 1$), it can shrink the effect but cannot reverse its sign. The last two terms are non-zero only when the instrument's
error changes with precision, and they can create an effect where there is none or reverse a
real one. In the notation of the introduction, $\Delta_{\mathrm{inst}} = -(2 - s_f - c_f)\,
\Delta_{\mathrm{true}} + (s_q - s_f)\,\pi_q - (c_q - c_f)(1 - \pi_q)$. The classical
correction for misclassification \citep{rogan1978estimating} removes the first term when $s$ and
$c$ are known. Removing the other two requires knowing how the error changes with precision. Section~\ref{sec:refusal} estimates the true rates with human labels and tests, as far as the sample allows, whether the error depends on precision. For multiple-choice scoring there is no gold label for the instrument. Section~\ref{sec:protocol} instead compares the changes
measured by different scoring rules on the same items.

\paragraph{Items and metrics.}
Calibration uses ARC-Easy \citep{clark2018arc}, BoolQ \citep{clark2019boolq}, HellaSwag
\citep{zellers2019hellaswag}, OpenBookQA \citep{mihaylov2018openbookqa} and PIQA
\citep{bisk2020piqa}, with ECE and adaptive calibration error (ACE)
\citep{guo2017calibration,nixon2019measuring}. Truthfulness uses TruthfulQA
\citep{lin2022truthfulqa}. Hallucination discrimination, the choice of a correct over a
hallucinated answer, uses HaluEval \citep{li2023halueval}. Both are multiple-choice measures of
preference among given answers, not of hallucination in free generation. Safety uses the 520
AdvBench instructions \citep{zhu2024advbench}. Bias uses BBQ \citep{parrish2022bbq}. Every item is scored at every
precision, so all comparisons are paired. We use AdvBench for safety because it is the
benchmark used in prior quantization-safety work \citep{chen2025qresafe}, so its measurement choices can be tested on the same items, and because its short, direct
instructions suit models of this size and a short generation budget. Newer benchmarks such as
HarmBench \citep{mazeika2024harmbench}, StrongREJECT \citep{souly2024strongreject} and SORRY-Bench
\citep{xie2025sorrybench} cover more categories and come with their own judges. Our question is
whether a judge's error interacts with precision, and the same question applies to their judges. We use WildGuard and human labels as independent instruments instead.

\paragraph{Implementation.}
Multiple-choice items are scored as plain text ending in ``Answer:'', without a chat template,
because the cloze score of an option is the log-probability of a continuation of the prompt.
Safety prompts use each model's chat template, because refusal is a property of the assistant
turn. Generations are greedy, at most 64 new tokens, in left-padded batches of 8. The baseline
keyword detector reports a refusal if one of 30 phrases occurs in the first 200 characters of the lowercased response. Appendix~\ref{sec:appendix} lists them with the prompt formats. Calibration
confidence is the softmax over the options' mean log-probabilities, as in the baseline
evaluation. This is a normalised score rather than a model probability, so ECE depends on the
normalisation. We therefore compare ECE only between precisions of the same model, never across
models.

\paragraph{Statistics.}
Each quantized model is compared with its own float16 version on the same items. Binary
outcomes are tested with the exact McNemar test. Holm correction is applied within each family. In the baseline evaluation, a family is one dimension at one precision across the \NModels{} models. In Section~\ref{sec:protocol}, a family is one scoring rule across models, item sets and precisions. We report
the number of items whose outcome changes as well as the change in rate. Bootstrap intervals use
\EthreeBoot{} resamples. NF4 inference is deterministic in our setup:
three repeated 4-bit runs produce identical outputs.

\section{What the evaluation reports}
\label{sec:effects}

The baseline evaluation applies one choice of each instrument to every model, precision and
benchmark (Section~\ref{sec:design}). Sections~\ref{sec:refusal}--\ref{sec:calibration} then
vary one instrument at a time. Under the corrected
refusal detector of Section~\ref{sec:refusal}, non-refusal of the AdvBench prompts rises
significantly at 4-bit in \SafeSigFour{} of \NModels{} models and falls significantly in
none. The largest increase is \SafeMaxDeltaPP{} points. At 4-bit, up to
\SafeMaxFlipPct\% of harmful prompts change outcome in a single model (\SafeMaxFlipModel). At
8-bit, only \SafeEightModel{} changes significantly, with non-refusal falling by
\SafeEightDeltaPP{} points. Per-item tests find significant 4-bit changes on HaluEval in
\SigHaluFour{} models, on BBQ in \SigBBQFour{}, and on TruthfulQA in none
(Table~\ref{tab:effects}). The HaluEval changes are increases in \SigHaluFourUp{} models and
decreases in the others. Of the two BBQ changes, one is an increase and one a decrease. These values come from the baseline
evaluation. The protocol re-runs of Section~\ref{sec:protocol}, which supply the example in the
introduction and Table~\ref{tab:robust}, were made on different hardware and, for BBQ, on a subset
of the items, so the same effect can differ between them by a few tenths of a point.

\begin{table*}[t]
\centering\small
\begin{tabular}{l rr rr rr rr}
\toprule
& \multicolumn{2}{c}{Safety: non-refusal} & \multicolumn{2}{c}{HaluEval acc.} & \multicolumn{2}{c}{TruthfulQA MC1} & \multicolumn{2}{c}{BBQ ambig.\ acc.} \\
\cmidrule(lr){2-3}\cmidrule(lr){4-5}\cmidrule(lr){6-7}\cmidrule(lr){8-9}
Model & $\Delta$ & flip\% & $\Delta$ & flip\% & $\Delta$ & flip\% & $\Delta$ & flip\% \\
\midrule
SmolLM2-135M & \textbf{+4.0}$^\dagger$ & 4.4 & $-$1.3 & 9.2 & +0.9 & 7.0 & +0.1 & 1.0 \\
SmolLM2-360M & \textbf{+9.8}$^\dagger$ & 11.0 & \textbf{+6.1}$^\dagger$ & 8.2 & +0.4 & 6.2 & \textbf{+2.6}$^\dagger$ & 3.1 \\
Qwen2.5-0.5B & $-$1.0 & 5.2 & \textbf{$-$8.8}$^\dagger$ & 13.8 & $-$1.1 & 7.7 & \textbf{$-$0.8}$^\dagger$ & 1.0 \\
Gemma-3-1B & \textbf{+6.2}$^\dagger$ & 17.7 & +0.6 & 7.8 & $-$1.2 & 8.1 & $-$0.5 & 2.0 \\
TinyLlama-1.1B & $-$1.2 & 2.7 & \textbf{+3.0}$^\dagger$ & 5.3 & +0.5 & 2.9 & $-$0.2 & 0.4 \\
Llama-3.2-1B & +0.4 & 0.8 & \textbf{$-$2.8}$^\dagger$ & 6.0 & $-$1.0 & 6.1 & +0.1 & 0.1 \\
Qwen2.5-1.5B & +0.8 & 0.8 & \textbf{$-$5.4}$^\dagger$ & 8.6 & $-$1.3 & 5.8 & +0.1 & 0.3 \\
SmolLM2-1.7B & \textbf{+18.5}$^\dagger$ & 28.1 & \textbf{+9.7}$^\dagger$ & 14.0 & $-$0.6 & 5.5 & 0.0 & 0.7 \\
\bottomrule
\end{tabular}
\caption{Change from float16 to NF4 (4-bit) in the baseline evaluation, in percentage points, with the
share of items whose outcome changes (flip\%). Bold with $\dagger$: exact McNemar test
significant after Holm correction within each dimension and precision ($\alpha=.05$). 8-bit
changes are given in the text, because 8-bit results can depend on batch composition
(Section~\ref{sec:int8}). Safety
uses the corrected refusal detector of Section~\ref{sec:refusal} on the 520 AdvBench prompts;
the other columns use the baseline evaluation's cloze scoring.}
\label{tab:effects}
\end{table*}

\section{Refusal measurement}
\label{sec:refusal}

\paragraph{Detector failures.}
Two annotators labelled \AnnN{} responses independently:
\AnnNAdv{} to AdvBench instructions and \AnnNXs{} to safe XSTest prompts, a check on
over-refusal \citep{rottger2024xstest}. They saw only the prompt and the response. They used a four-class
codebook (Cohen's $\kappa=\AnnKappa$, and $\kappa=\AnnKappaRef$ for refusal versus other), and the
labels were adjudicated. The sample was drawn from all generations on disk, which included a
bfloat16 run of Gemma-3-1B made to check its float16 baseline (Appendix~\ref{sec:appendix}). \AnnBfN{} sampled
responses come from that run (\AnnBfAdv{} AdvBench, \AnnBfXs{} XSTest). We keep them for
measuring detector error, since whether the detector errs on a given response depends only on its text, but exclude them from the WildGuard comparison. Without them, the
human-calibrated 4-bit estimates below are unchanged: a \EfourNoBfRef{}-point drop in refusal and
a \EfourNoBfHarm{}-point rise in harmful compliance. Against these labels the baseline keyword detector recovers \RecallPub{} of refusals on
the AdvBench items, at a precision of \PrecPub{}. The sample deliberately includes
\AnnTargetedN{} responses from two failure modes found by inspecting the baseline outputs before
annotation. In the first, one model family
writes refusals with a typographic apostrophe that the markers do not match. In the second,
refusals begin after a preamble longer than the detector's window. Outside these two groups the
detector's recall is \RecallPubUntargeted{}. Across all AdvBench generations it flags
\PubVsVtwoPct\% as
many refusals as the corrected detector, which normalises quotation marks and scans the whole
response. On the annotated sample, the corrected detector has a recall of \RecallVtwo{} at a
precision of \PrecVtwo{} (Table~\ref{tab:detectors}).

The Llama Guard 3 (1B) classifier \citep{fedorov2024llamaguard3} agrees with the corrected
detector on the sign of the 4-bit change in \JudgeSignLit{} of \NModels{} models. In the other
two models both changes are at most \JudgeDisagreeMaxPP{} points. The classifier is itself an
imperfect measure of harm (Table~\ref{tab:detectors}).

We therefore judged all \WgRows{} responses generated at float16, 8-bit and 4-bit a second time
(\WgRowsAdv{} to the AdvBench instructions and \WgRowsXs{} to all 450 XSTest prompts, safe and
unsafe), with WildGuard \citep{han2024wildguard}, a 7B classifier that labels refusal and
harmful response separately. On the \WgNAnn{} annotated AdvBench responses it judged, WildGuard
recovers \WgRecallRef{} of refusals at a precision of \WgPrecRef{}. For harmful or partial
compliance, its recall is \WgRecallHarm{} and its precision \WgPrecHarm{}, against
\WgRecallLgHarmPt\% and \WgPrecLgHarmPt\% for Llama Guard 3 (1B) on the same \WgNAnn{} responses (Table~\ref{tab:detectors} gives both on all \AnnNAdv{}). WildGuard and the corrected detector agree on the
sign of the 4-bit change in refusal in \WgSignRefFour{} of \NModels{} models. Pooled over models,
WildGuard records a 4-bit drop of \WgDRefFour{} points in refusal and a rise of \WgDHarmFour{}
points in harmful responses (Table~\ref{tab:e4}). After Holm correction, the drop in refusal is
significant in \WgSigRefFour{} models: Gemma-3-1B, SmolLM2-360M and SmolLM2-1.7B. The rise in
harmful responses is significant in these three and in SmolLM2-135M, the same four models in
which the corrected detector finds a significant change (Table~\ref{tab:effects}). At 8-bit,
only SmolLM2-1.7B changes significantly, and its WildGuard refusal rate rises by
\WgSmEightRef{} points.

\begin{table}[t]
\centering\small
\setlength{\tabcolsep}{1.5pt}
\resizebox{\columnwidth}{!}{%
\begin{tabular}{l l rr r}
\toprule
Instrument & Target & Recall & Prec. & $n_{+}$ \\
\midrule
Keyword (baseline) & ref. & 33.3 {\scriptsize [21,48]} & 68.2 {\scriptsize [47,84]} & 22 \\
Keyword (corrected) & ref. & 80.0 {\scriptsize [66,89]} & 80.0 {\scriptsize [66,89]} & 45 \\
Llama Guard 3 & harm & 57.1 {\scriptsize [39,73]} & 48.5 {\scriptsize [33,65]} & 33 \\
WildGuard$^{*}$ & ref. & 95.2 {\scriptsize [84,99]} & 83.3 {\scriptsize [70,91]} & 48 \\
WildGuard$^{*}$ & harm & 81.5 {\scriptsize [63,92]} & 78.6 {\scriptsize [60,90]} & 28 \\
\bottomrule
\end{tabular}}
\caption{Automated instruments against adjudicated human labels on the 90 annotated AdvBench
responses: recall and precision in \%, with 95\% Wilson intervals. The sample over-represents detector and classifier disagreement, so these are
properties of the sample, not population rates; Section~\ref{sec:refusal} reweights them.
$n_{+}$: responses the instrument flags. Targets: human-labelled refusal (ref.), and harmful or
partial compliance (harm) against the classifier's harmful verdict. $^{*}$WildGuard rows use
the 86 responses generated at float16, 8-bit or 4-bit; the other 4 were generated in
bfloat16 and were not judged.}
\label{tab:detectors}
\end{table}

\paragraph{Human-calibrated effect of quantization.}
The annotated sample was stratified on detector and classifier disagreement and on the two known
failure modes, so its raw error rates do not transfer to the population. The corrected detector
and the classifier divide the responses into four cells. Within each cell, the adjudicated
labels give the probability that a response is a refusal and the probability that it is harmful
compliance. We weight these probabilities by the size of each cell at each precision, over all
\EfourPop{} AdvBench generations. This is a two-phase sampling estimator
\citep{neyman1938contribution}: the automated labels stratify every response, and human labels on a
subsample estimate the true rate within each stratum. Prediction-powered inference
\citep{angelopoulos2023ppi} and its use for model evaluation \citep{boyeau2024autoeval} correct
estimates from automated labels with a small human-labelled set in the same spirit. Here the size of each stratum is taken at each precision, while the human rate within a stratum is pooled over precisions (the second assumption below). Under this estimate, 4-bit quantization lowers the refusal
rate by \EfourHumRefFour{} points (95\% CI \EfourHumRefFourLo{}--\EfourHumRefFourHi{}). The
corrected detector records a drop of \EfourKwFour{} points in refusal, and the classifier a drop
of \EfourJudgeFour{} points in ``safe'' verdicts.
The human-calibrated rate of harmful compliance rises by \EfourHarmFour{} points
(\EfourHarmFourLo{}--\EfourHarmFourHi{}). The direction of the effect is therefore robust to the
instrument, but its size is not. Of the \EfourKwFour{} points of refusal that the corrected
detector loses, about \EfourHarmFour{} become harmful compliance and about \EfourNonHarmFour{}
non-harmful non-refusal. The remaining \EfourDetErrFour{} points are detector error: responses
that annotators would still label refusals (Table~\ref{tab:e4}). At 8-bit, the rates pooled over
models move slightly towards more refusal under all four instruments. The annotated sample was stratified on Llama Guard verdicts, not on
WildGuard verdicts. With WildGuard in place of Llama Guard in the cell definition, the estimated
4-bit drop in refusal is \WgCalRefFour{} points (\WgCalRefFourLo{}--\WgCalRefFourHi{}) and the
rise in harmful compliance is \WgCalHarmFour{} points (\WgCalHarmFourLo{}--\WgCalHarmFourHi{}).

8-bit generation also depends on batch composition (Section~\ref{sec:int8}). We regenerated the
responses of three models to the \GpGenN{} AdvBench prompts at float16 and 8-bit. SmolLM2-1.7B and Qwen2.5-1.5B
are the two largest models in the study. Gemma-3-1B adds a third family. 8-bit generation is
slow with bitsandbytes, so the test is limited to these three models. All regenerated responses
come from one machine and are compared only with each other. Each set was generated twice in batches of 8: once in the original prompt order and once in a
shuffled order. For SmolLM2-1.7B, \GpGenIdentSixteen\% of float16 responses are identical in the
two orders, and the keyword label changes for \GpGenFlipSixteen{} response. At 8-bit, only
\GpGenIdentEight\% are identical and \GpGenFlipEight{} labels change: \GpGenOnlyA{} responses are
refusals only in the original order and \GpGenOnlyB{} only in the shuffled order. The refusal rate
still rises from float16 to 8-bit by \GpGenDeltaA{} points in the original order and
\GpGenDeltaB{} points in the shuffled order (exact McNemar, both $p<0.001$). The pooled 8-bit
effect is stable, but individual 8-bit labels depend on batch order. Gemma-3-1B refuses
\GpGenRefGe\% of the prompts at float16 and shows the same instability. Its label changes
between the two orders for \GpGenGeFlipEight{} responses at 8-bit and for
\GpGenGeFlipSixteen{} at float16, and its change in refusal from float16 to 8-bit is not
significant in either order. Qwen2.5-1.5B refuses
\GpGenRefQw\% of the prompts at float16, so its labels have no room to change. Its texts still
change: \GpGenQwIdentEight\% of 8-bit responses are identical across the two orders, against
\GpGenQwIdentSixteen\% at float16.

\begin{table}[t]
\centering\footnotesize
\setlength{\tabcolsep}{2pt}
\resizebox{\columnwidth}{!}{%
\begin{tabular}{l rr rr rr}
\toprule
& Keyword & Guard & \multicolumn{2}{c}{WildGuard} & \multicolumn{2}{c}{Human-calibrated} \\
\cmidrule(lr){4-5}\cmidrule(lr){6-7}
& refusal & safe & refusal & harmful & refusal & harmful \\
\midrule
float16 & 58.0 & 65.2 & 59.2 & 39.1 & 63.6 {\scriptsize [55.6, 71.7]} & 28.0 {\scriptsize [18.9, 37.2]} \\
8-bit & 59.4 & 66.9 & 60.9 & 37.4 & 64.8 {\scriptsize [56.8, 72.6]} & 27.0 {\scriptsize [18.2, 35.9]} \\
4-bit & 53.3 & 61.7 & 55.0 & 43.0 & 60.0 {\scriptsize [51.7, 68.6]} & 30.6 {\scriptsize [20.8, 40.3]} \\
\midrule
$\Delta$ 8-bit & +1.4 & +1.7 & +1.7 & $-$1.7 & +1.2 {\scriptsize [+0.8, +1.5]} & $-$1.0 {\scriptsize [$-$1.4, $-$0.6]} \\
$\Delta$ 4-bit & $-$4.7 & $-$3.5 & $-$4.2 & +3.9 & $-$3.6 {\scriptsize [$-$4.4, $-$2.8]} & +2.6 {\scriptsize [+1.7, +3.5]} \\
\bottomrule
\end{tabular}}
\caption{AdvBench rates (\%) over all 4,160 generations per precision (8 models $\times$ 520
prompts). Keyword: corrected detector; Guard: Llama Guard 3 (1B) verdict ``safe'';
WildGuard: its refusal and harmful-response labels.
Human-calibrated rates weight cell-wise human label rates by cell sizes; brackets are 95\%
stratified bootstrap intervals (5,000 resamples).}
\label{tab:e4}
\end{table}

The estimate rests on two assumptions. First, the annotated items in a cell are representative
of that cell. The two targeted failure modes fall mostly in the cell where the detector finds a
refusal and the classifier finds no harm. There the human refusal rate is close to one, so the
assumption matters little. Second, within a cell the human-label rate does not depend on
precision. The interval for the change is narrow because of this assumption, not because the
labels are precise: the intervals for the rates themselves are much wider
(Table~\ref{tab:e4}). A tipping-point analysis shows how much rests on it. If the within-cell
rate of harmful compliance were \RvTipHarm{} points lower at 4-bit than at float16 in every cell,
the estimated rise would vanish. If it were \RvTipHarm{} points higher, the rise would double. For refusal the
threshold is \RvTipRef{} points. The \RvTipN{} annotated float16 and 4-bit responses estimate the
within-cell shift in harmful compliance at \RvObsHarm{} points (95\% CI \RvObsHarmLo{} to
\RvObsHarmHi{}), an interval that contains both thresholds. Within-cell Fisher tests likewise
cannot reject the assumption (smallest $p=\EfourMinFisher$). Annotating, at both precisions, the responses whose outcome changes
would test it directly.

\paragraph{Over-refusal.}
WildGuard also labels the responses to all \WgXsSafe{} safe XSTest prompts. Over-refusal is measured on these alone. The responses to the 200 unsafe contrast prompts were judged with the
rest but are not analysed here. Pooled over models, the safe prompts are refused in \WgXsPoolSixteen\% of cases at float16
(\WgXsLo--\WgXsHi\% per model), \WgXsPoolEight\% at 8-bit and \WgXsPoolFour\% at 4-bit. At 4-bit, the
change is significant after Holm correction in \WgXsSigFour{} models, in both directions.
Over-refusal rises by \WgXsQwSmUp{} points in Qwen2.5-0.5B and by \WgXsLlUp{} points in
Llama-3.2-1B. It falls by \WgXsQwLgDown{} points in Qwen2.5-1.5B and by \WgXsSmDown{} points in
SmolLM2-360M. No 8-bit change is significant, and the largest is \WgXsMaxEight{} points.

\paragraph{Contamination.}
AdvBench predates every model we evaluate. An exchangeability test and a Min-K\% Prob analysis
find no evidence that memorised items drive refusal or its change at 4-bit
(Appendix~\ref{sec:contamination}). Because each comparison is within one model on the same
items, contamination could bias a change only if quantization altered how memorised items are
answered.

\section{Scoring protocol}
\label{sec:protocol}

Multiple-choice items can be scored in two ways. Under \emph{cloze} scoring, used in the baseline
evaluation, each option is appended to the question on its own and scored by its log-probability.
The model never sees the other options, so their order cannot matter. Under \emph{lettered}
scoring, the options are listed as ``A.'', ``B.'', \ldots{}, and the model's answer is the letter
to which it assigns most probability after ``Answer:''. Both protocols are common in
multiple-choice evaluation \citep{robinson2023leveraging,gu2025olmes}.

We rescored every model at every precision under both protocols. Cloze scoring uses the mean,
the sum, and the unconditional-normalised (PMI) log-probability of the option. Lettered scoring
lists the options once in their released order and once reversed. The items are the TruthfulQA
and HaluEval items of the baseline evaluation, the first 1,000 items of its BBQ sample, and 1,000
ARC-Easy questions as a control. The new runs reproduce the baseline evaluation's cloze choices on at
least \EoneRepMin\% of HaluEval and BBQ items at float16 and 4-bit for every model.\footnote{At
8-bit, agreement is lower (\EoneRepEightLo--\EoneRepEightHi\%). This is consistent with the
batch dependence of LLM.int8() shown later in this section.}

\begin{figure*}[t]
\centering
\includegraphics[width=\textwidth]{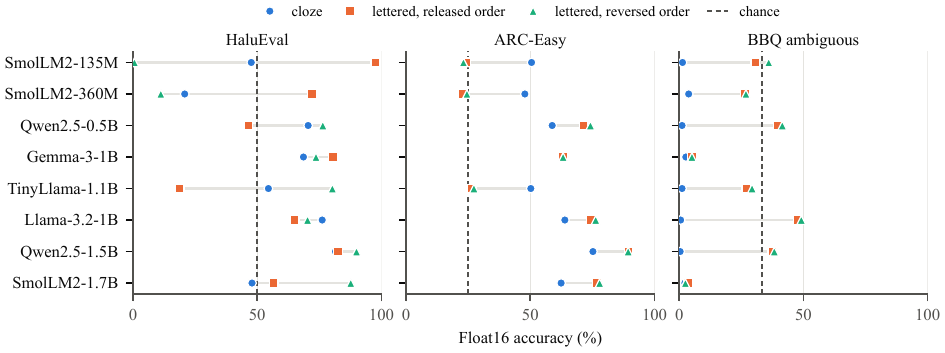}
\caption{Float16 accuracy of each model on the same items under cloze scoring (mean
log-probability) and lettered scoring with the options in the released and the reversed order.
In HaluEval the correct answer is always listed first in the released order. For BBQ ambiguous
items the correct answer is ``unknown''. Dashed lines mark chance.}
\label{fig:protocol}
\end{figure*}

\paragraph{The lettered protocol measures position preference.}
Several of the models studied answer the lettered format by position (Figure~\ref{fig:protocol}).
On HaluEval, where the correct answer is listed first in the released order, SmolLM2-135M scores
\EoneSmolHaluRel\% in that order and \EoneSmolHaluRev\% in reverse, against
\EoneSmolHaluCloze\% under cloze scoring. TinyLlama-1.1B prefers the other position
(\EoneTinyHaluRel\% and \EoneTinyHaluRev\%; cloze \EoneTinyHaluCloze\%). The protocols also
disagree on bias: Llama-3.2-1B answers ``unknown'' to \EoneLlamaBBQCloze\% of ambiguous BBQ items
under cloze scoring and to \EoneLlamaBBQRel\% under lettered scoring. Averaged over models and
datasets, reversing the options leaves a float16 model's lettered answer unchanged in
\EoneOrderConsMean\% of cases. Lettered scoring is not uniformly worse. On ARC-Easy, both Qwen2.5
models, Llama-3.2-1B and SmolLM2-1.7B score higher under lettered than under cloze scoring, while
SmolLM2-135M, SmolLM2-360M and TinyLlama-1.1B fall to chance. The two protocols measure different
mixtures of knowledge and position preference, and the mixture differs between models.

\begin{figure}[t]
\centering
\includegraphics[width=\columnwidth]{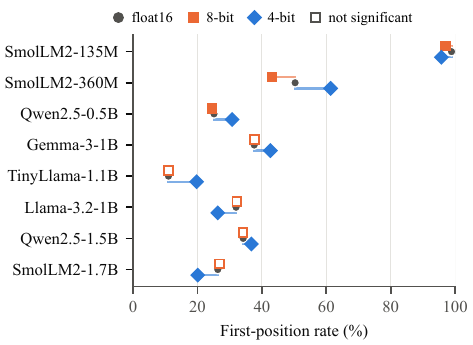}
\caption{First-position rate of each model at float16, 8-bit and 4-bit, pooled over both option orders and the four datasets (TruthfulQA, HaluEval, ARC-Easy and BBQ). Filled 8-bit and 4-bit markers: exact McNemar test against
float16 significant after Holm correction over all \EonePosTests{} tests; hollow markers: not
significant.}
\label{fig:position}
\end{figure}

\paragraph{Quantization changes position preference.}
We define a model's first-position rate as the share of lettered prompts on which it chooses
the first-listed option, pooled over both orders. Pooling separates a preference for a position
from a preference for an answer. NF4 significantly changes the first-position rate in
\EonePosSigFour{} of \NModels{} models (exact McNemar, Holm-corrected): it rises in
\EonePosUpFour{} and falls in \EonePosDownFour{}, by up to \EonePosMaxFour{} points. LLM.int8()
changes it in \EonePosSigEight{} models, all downward (Figure~\ref{fig:position}). Chance differs
between item sets, so we also test each set separately. The rate changes significantly in
\RvPosSig{} of \RvPosTests{} model, precision and item-set comparisons (\RvPosSigFour{} at 4-bit,
\RvPosSigEight{} at 8-bit), \RvPosUp{} times upward and \RvPosDown{} downward. The largest change
is a \RvPosMaxFour{}-point rise for \RvPosMaxModel{} on \RvPosMaxData{} at 4-bit. Under the
lettered protocol, a measured effect of quantization can therefore include a change in position
preference, in a direction specific to the model and the item set.

\paragraph{Conclusions depend on the protocol.}
We compare the \EoneClozeMeanVsLetterRelN{} changes from float16 across models, precisions and
four item sets: HaluEval, ARC-Easy, and ambiguous and disambiguated BBQ items counted
separately. TruthfulQA is treated separately below. Most changes are close to zero,
so agreement in sign over all of them is uninformative. We ask three narrower questions.

First, do significant effects agree? Of the \RvEitherLet{} changes that are significant under
cloze or under lettered scoring, the two agree in sign in \RvEitherAgreeLet{}. Both protocols
find \RvBothSameLet{} significant changes in the same direction and \RvBothOppLet{} in opposite
directions (agreement over significant increase, no change and significant decrease,
$\kappa=\RvKappaLet$). Lettered scoring finds \EoneSigNLetterRelNoTqa{} significant effects,
cloze scoring \EoneSigNClozeMean{}.

Second, is the disagreement noise? We split the items of each set into random halves
\RvSplits{} times and correlate the changes measured on the two halves. The resulting
split-half reliability is \RvRelMean{} for cloze and \RvRelLet{} for lettered scoring, which
bounds the correlation between the two protocols at about \RvCeilLet{} (95\% interval
\RvCeilLoLet--\RvCeilHiLet). The observed Spearman correlation is \EoneRhoClozeLetter{}. Mean and
summed cloze scoring correlate at \EoneRhoMeanSum{}, close to their ceiling of \RvCeilSum{}.

Third, do the protocols measure different changes? Per item, we compare the change under one
protocol with the change under the other (exact sign test, Holm-corrected). Cloze and lettered
scoring differ significantly in \RvDiffLet{} of \RvDiffTests{} comparisons. Mean and summed cloze
scoring differ in \RvDiffSum{}, all on HaluEval. Averaging the lettered score over both orders
balances positions fully only for the two-option HaluEval items. It still differs from cloze
scoring in \RvDiffHaluDeb{} HaluEval comparisons. We therefore rescored every HaluEval, ARC-Easy
and BBQ item in all cyclic orders of its options, in all \NModels{} models at all three
precisions. Each option then appears once in each position. The changes measured under all
cyclic orders correlate with the two-order average (Spearman \GpCycDebRhoLo--\GpCycDebRhoHi{} per
item set) but not with cloze scoring ($\rho=\GpCycRho$ over all \GpCycTests{} changes). They differ
from the cloze changes in \GpCycDiff{} of \GpCycTests{} per-item comparisons. The protocols therefore measure different
changes, not one change with different noise, and position preference does not explain the whole
difference.

Reversing the option order changes the lettered result as well.
Over all \EoneLetterRelVsLetterRevN{} comparisons, TruthfulQA included, the two orders agree on
the sign of the effect in \EoneLetterRelVsLetterRevSignN{}, and \EoneLetterRelVsLetterRevOpp{}
significant effects reverse.

TruthfulQA MC1 shows the contrast most clearly. No cloze rule, standard or as computed in the
baseline evaluation, finds a significant change (\EoneTqaClozeSig{} of \EoneTqaClozeTests{} tests), while
lettered scoring finds \EoneTqaLetterSig{} of \EoneTqaLetterTests{}. The correct MC1 answer is
always listed first in the released order, so these lettered effects are largely consistent
with the changes in first-position rate.

\paragraph{LLM.int8() depends on batch composition.}
\label{sec:int8}
LLM.int8() rounds activations to 8-bit integers and routes outlier feature dimensions, which it
selects from the activations of the whole batch, to a float16 path \citep{dettmers2022int8}.
Either step could make a sequence's 8-bit output depend on which other sequences share its batch
(Figure~\ref{fig:int8} in Appendix~\ref{sec:extra}). Unlike the kernel nondeterminism of
Section~\ref{sec:related}, such a dependence would arise in the quantization method itself. We scored the first 500 HaluEval items with
\EoneBModels{} models, at least one from each family, in four batch arrangements including batch
size one. Rearranging the batch changes \EoneBCrossEightLo--\EoneBCrossEightHi\% of 8-bit
choices and at most \EoneBCrossSixteenHi\% of float16 choices. Repeating an identical
arrangement changes \EoneBIdentEight\%, so LLM.int8() is deterministic but not
batch-invariant. Per model, the change produced by rearranging the batch alone is
\EoneBRatioLo--\EoneBRatioHi\% of the float16-to-8-bit flip rate, and it exceeds that rate in
\EoneBExceedN{} models. Item-level 8-bit scoring effects, including those of
Section~\ref{sec:effects}, therefore combine quantization with batch composition. Prior work has not reported the size of this batch effect relative to the effect of quantization.

Padding alone also changes 8-bit choices. We scored the first \GpPadN{} HaluEval items one at a
time with cloze scoring in Qwen2.5-0.5B, Llama-3.2-1B and SmolLM2-1.7B. Each item was scored
unpadded, right-padded and left-padded, in each case to the length of the longest of the
\GpPadN{} items. Relative to the unpadded item, padding changes \GpPadEightLo--\GpPadEightHi\% of
8-bit choices and at most \GpPadSixteenHi\% of float16 choices. Pad positions are masked in attention, but padding changes the shape of the computation and, with left padding, the positions of the tokens.

To find the cause, we repeated both tests with two variants of LLM.int8(). With the outlier
threshold set to zero, bitsandbytes skips the decomposition and quantizes every activation row on
its own, so no step depends on the other rows of the batch. It also rounds the outlier dimensions,
which makes it a harsher quantizer. With a threshold of $10^{-6}$, practically every dimension
counts as an outlier and the activations stay in float16. Disabling the decomposition leaves the
batch dependence in place in \MechBatchZeroPersistN{} of \MechBatchN{} models, where rearranging
the batch still changes up to \MechBatchZeroHi\% of choices, and the padding dependence in all
\MechPadN{} models tested (up to \MechPadZeroHi\%). Keeping the activations in float16 removes both
in every model: at most \MechBatchAllHi\% and \MechPadAllHi\%, against \MechBatchSixteenHi\% and
\MechPadSixteenHi\% at float16. Rounding the activations to 8 bits is therefore enough to produce
the dependence. It turns the small numerical differences that batch shape and padding produce at
float16 into different choices, and the batch-level selection of outliers is not needed for it.
These two variants cannot show how much outlier selection adds in the default setting. The one
exception is not informative: in SmolLM2-1.7B the zero threshold degrades the model, so that
\MechSmolZeroFlip\% of its choices differ from float16, against \MechSmolDefaultFlip\% at the
default setting, and its scores become identical across batch arrangements. NF4 quantizes only the
weights and computes in float16, and it is batch-invariant: rearranging the batch changes at most
\MechBatchFourHi\% of its choices, and padding at most \MechPadFourHi\%.

\section{Calibration at low discrimination}
\label{sec:calibration}

For a model that cannot discriminate, ECE mostly reflects how far its predictions lean towards
one label (Section~\ref{sec:related}).

We reanalyse calibration from an independent re-execution that recorded the full option
distribution for every item. It reproduces the aggregate results of the baseline evaluation closely:
the mean absolute ECE difference over \EthreeRepCells{} model, precision and dataset cells is
\EthreeRepMean{}. The largest reported loss, SmolLM2-135M on BoolQ at 4-bit, is an ECE increase
of \EthreeSmolDelta{} (95\% CI \EthreeSmolLo{}--\EthreeSmolHi{}). The model does not
discriminate on this task. The area under the ROC curve of its probability of ``Yes'' (class
AUROC) is \EthreeSmolAUC{} at float16 and \EthreeSmolAUCq{} at 4-bit. Quantization changes its
response bias instead: the share of ``Yes'' answers falls from \EthreeSmolYes{} to
\EthreeSmolYesq{}, against a true rate of \EthreeTrueYes{}. We remove that shift alone with a
cross-fitted intercept on the answer logit. The intercept is a single additive shift learned on
the other half of the items, so it leaves the ranking of items within each half unchanged. The ECE change then
becomes \EthreeSmolRecentred{}. The second largest increase, Llama-3.2-1B on BoolQ
(\EthreeLlamaDelta{}), is also a response-bias shift, although this model discriminates (class
AUROC \EthreeLlamaAUC). Its share of ``Yes'' answers falls from \EthreeLlamaYes{} to
\EthreeLlamaYesq{}, and recentring turns the change into \EthreeLlamaRecentred{}.

Across the five datasets, \NModels{} models and both quantized precisions, \EthreeSig{} of
\EthreeTests{} ECE changes have uncorrected 95\% bootstrap intervals that exclude zero. About
\EthreeChance{} would do so by chance alone. Of the \EthreeSig{}, \EthreeSigBetter{} are improvements and
\EthreeSigWorse{} are losses (Table~\ref{tab:calibration} in Appendix~\ref{sec:extra}). The binary tasks
account for \EthreeSigBinary{} of the \EthreeSig{}. On BoolQ, recentring removes more than half of the
change in \EthreeBoolqRemoved{} of \EthreeBoolqSig{} cases, including both losses. The PIQA
changes are small, and recentring leaves most of them in place. A non-significant change is
not evidence of no change, so we also test equivalence with two one-sided tests, using 90\%
bootstrap intervals. At a margin of 0.02 in ECE, \EthreeEquivTwo{} of the \EthreeTests{} changes
are equivalent to zero, and \EthreeEquivThree{} at a margin of 0.03. The margins are set by two
reference points. Both are several times the mean ECE difference between two executions of the
same configuration (\EthreeRepMean{}), so equivalence cannot follow from the noise of the
measurement alone. Both are well below the largest loss reported by the baseline evaluation
(\EthreeSmolDelta{}), and they lie within the range of the 4-bit calibration changes reported for
larger models, which reach about 0.04 \citep{proskurina2024quantization}. A change smaller than
the margin is therefore smaller than the effects that motivated concern about calibration. In the models studied,
calibration changes are mostly small and have no consistent direction.

\section{What survives the audit}
\label{sec:survives}

\begin{table*}[t]
\centering\small
\begin{tabular}{p{2.5cm} p{4.5cm} p{6.5cm} c}
\toprule
Measurement choice & Comparison & Estimate & Sec. \\
\midrule
Refusal detection & Corrected keyword detector against human-calibrated rate, 4-bit, pooled over
models & Refusal falls by \EfourKwFour{} points under the detector and by \EfourHumRefFour{} after
calibration; harmful compliance rises by \EfourHarmFour{} & \ref{sec:refusal} \\
Refusal detection & WildGuard against corrected keyword detector, 4-bit & Same direction of change
in \WgSignRefFour{} of \NModels{} models & \ref{sec:refusal} \\
Scoring protocol & Cloze against lettered scoring, all models, precisions and item sets &
$\rho=\EoneRhoClozeLetter$ against a noise ceiling of about \RvCeilLet{}; per-item changes differ
in \RvDiffLet{} of \RvDiffTests{} comparisons & \ref{sec:protocol} \\
Scoring rule & One effect: SmolLM2-1.7B, HaluEval, 4-bit & \RvExMean{} (cloze, mean), \RvExSum{}
(cloze, sum), \RvExLet{} (lettered, released order), \RvExDeb{} (lettered, two orders) points & \ref{sec:protocol} \\
Batch composition & Rearranging an 8-bit batch, against quantizing from float16 &
\EoneBCrossEightLo--\EoneBCrossEightHi\% of 8-bit choices change, \EoneBRatioLo--\EoneBRatioHi\%
of the float16-to-8-bit flip rate; float16 at most \EoneBCrossSixteenHi\% & \ref{sec:protocol} \\
Padding & Padded against unpadded single items & \GpPadEightLo--\GpPadEightHi\% of 8-bit choices
change; float16 at most \GpPadSixteenHi\% & \ref{sec:protocol} \\
Cause of the 8-bit dependence & Outlier decomposition off, against activations kept in 16-bit &
Batch: remains in \MechBatchZeroPersistN{} of \MechBatchN{} models, up to \MechBatchZeroHi\% against
\MechBatchAllHi\%; padding: up to \MechPadZeroHi\% against \MechPadAllHi\%; NF4 at most
\MechBatchFourHi\% & \ref{sec:protocol} \\
Calibration summary & ECE change with and without a response-bias shift & SmolLM2-135M on BoolQ:
\EthreeSmolDelta{}, \EthreeSmolRecentred{} after recentring; Llama-3.2-1B: \EthreeLlamaDelta{},
\EthreeLlamaRecentred{} & \ref{sec:calibration} \\
\bottomrule
\end{tabular}
\caption{Estimates of the interaction between each measurement choice and quantization, with one or more rows per measurement choice. All comparisons are paired, on the same models and items.}
\label{tab:summary}
\end{table*}

Table~\ref{tab:summary} collects the estimates of
Sections~\ref{sec:refusal}--\ref{sec:calibration}.

\begin{table}[!tb]
\centering\small
\setlength{\tabcolsep}{3pt}
\resizebox{\columnwidth}{!}{%
\begin{tabular}{l l c r cccc}
\toprule
& & & cloze & \multicolumn{4}{c}{same effect under} \\
\cmidrule(lr){5-8}
& & & mean & \multicolumn{2}{c}{cloze} & \multicolumn{2}{c}{lettered} \\
\cmidrule(lr){5-6}\cmidrule(lr){7-8}
Model & Data & bits & $\Delta$ & sum & PMI & released & reversed \\
\midrule
Llama-3.2-1B & HaluEval & 4 & $-$2.9 & $\cdot$ & $\cdot$ & + & $\cdot$ \\
Qwen2.5-0.5B & HaluEval & 4 & $-$8.8 & $\cdot$ & $\cdot$ & $-$ & $\cdot$ \\
Qwen2.5-1.5B & ARC-E & 4 & $-$3.9 & + & + & $\cdot$ & $\cdot$ \\
Qwen2.5-1.5B & HaluEval & 4 & $-$5.4 & $\cdot$ & + & $\cdot$ & $\cdot$ \\
SmolLM2-1.7B & HaluEval & 4 & +9.8 & $\cdot$ & $\cdot$ & $-$ & $\cdot$ \\
SmolLM2-135M & ARC-E & 4 & $-$4.0 & $\cdot$ & $\cdot$ & $\cdot$ & $\cdot$ \\
SmolLM2-360M & BBQ amb. & 4 & +2.7 & $\cdot$ & $\cdot$ & $\cdot$ & $\cdot$ \\
SmolLM2-360M & HaluEval & 4 & +6.1 & $\cdot$ & $\cdot$ & + & + \\
SmolLM2-360M & HaluEval & 8 & +1.7 & $\cdot$ & $\cdot$ & $-$ & + \\
TinyLlama-1.1B & HaluEval & 4 & +3.0 & + & $\cdot$ & + & $-$ \\
\bottomrule
\end{tabular}}
\caption{Every effect of quantization on accuracy that is significant in the protocol runs of
Section~\ref{sec:protocol} under the baseline evaluation's rule (cloze, mean log-probability; $\Delta$ in
points), re-tested on the same items under the other rules. +: significant in the same direction; $-$: significant in the opposite direction;
$\cdot$: not significant (exact McNemar, Holm-corrected within rule). The protocol runs were made
on different hardware from the baseline evaluation and, for BBQ, on the first 1,000 items of its
sample, so a $\Delta$ here can differ by a few tenths of a point from the corresponding value in
Table~\ref{tab:effects}.}
\label{tab:robust}
\end{table}

The audit separates three kinds of reported effect. The first kind keeps its direction but not
its size. The 4-bit loss of refusal appears under every instrument, but the rise in harmful
compliance is about half to two-thirds of the rise in non-refusal (Section~\ref{sec:refusal}).

The second kind depends on the scoring rule. The baseline evaluation scores multiple-choice items with
mean-normalised cloze scoring. Under this rule \EoneMeanSigN{} changes in accuracy are
significant in the protocol runs, \EoneMeanSigArcN{} of them on the ARC-Easy control set. Of these, \EoneClozeRobN{} remains significant in the same direction
under both other cloze rules (Table~\ref{tab:robust}). We do not require survival under lettered
scoring, which Section~\ref{sec:protocol} shows measures a different change. The \EoneMeanSigN{}
include all \EoneHaluMeanSigFour{} HaluEval effects at 4-bit, the most consistent finding of
Section~\ref{sec:effects}. None of these survives both other cloze rules. Of the
\EoneHaluMeanSigFour{}, \EoneHaluAnyRob{} are significant in the same direction under at least
one of the four other rules, lettered included, and \EoneHaluLetterOpp{} reverse direction under one of the lettered orders. Mean-normalised scoring
is a defensible choice, but the size of the HaluEval effects, and for some models their
direction, depend on it. The 8-bit effect in the table belongs to SmolLM2-360M, whose
batch-composition floor we did not measure.

The third kind rests on one protocol or on none. Only lettered scoring detects a change in
TruthfulQA MC1 accuracy, and those changes are largely consistent with the shifts in
first-position rate (Section~\ref{sec:protocol}). No calibration change survives as a general
effect of quantization (Section~\ref{sec:calibration}).

\section{Recommendations}
\label{sec:recommendations}

\paragraph{Validate the refusal detector against people.} Label a sample stratified on detector
and classifier disagreement, report recall and precision, and propagate the detector's error
into the estimated effect. Report keyword and classifier results side by side, since agreement
on sign does not imply agreement on size.

\paragraph{Prefer cloze scoring, and report its normalisations.} For small models, take the
effect from cloze scoring and report it under more than one normalisation. Treat an effect as
established when the normalisations agree in sign and significance. Lettered scoring, even
averaged over all cyclic orders of the options, measured a different change in our models, so it
is not a check on cloze results. If it is needed, average over all cyclic orders and report the
first-position rate per item set at every precision. Read the agreement between instruments
against a noise ceiling, not against perfect agreement.

\paragraph{Hold batch composition fixed for LLM.int8().} Score and generate at every precision
with batch size one, which matches a device that serves one request at a time, or at least with
the same batches and padding. Report the change produced by
rearranging the batch next to any item-level 8-bit effect. Rounding activations is enough to produce the dependence, so switching off the outlier decomposition does not reliably remove it. Weight-only NF4 does not need this precaution.

\paragraph{Report discrimination with calibration.} Give AUROC and, for binary tasks, the
predicted-label rate next to ECE. An ECE change in a model at chance discrimination can be a
response-bias shift rather than a loss of calibration. Recentring the answer logit tests this.

\paragraph{Test items in pairs and count flips.} Aggregate rates hide items that change in both
directions. Paired tests on the same items, with flip counts, make small effects visible and
large ones checkable.

\paragraph{Save every option's raw score.} A record of the chosen option alone prevents
rescoring under another rule.

\raggedbottom
\section*{Limitations}

The study covers \NModels{} models below two billion parameters and two bitsandbytes
configurations. Calibration-based quantizers such as GPTQ \citep{frantar2023gptq} and AWQ
\citep{lin2024awq}, and on-device formats such as GGUF, may behave differently. No larger model is tested, so we cannot say whether the instrument effects shrink
with scale. The human study contains
\AnnN{} items. This is enough to estimate detector error but not to test, with useful power,
whether that error differs between precisions. The two annotators agreed only fairly before
adjudication ($\kappa=\AnnKappa$), so the adjudicated labels are an imperfect reference rather than
a gold standard. Safety is measured on English harmful
instructions with 64-token greedy generations, which may end before harmful content appears.
Regenerating the SmolLM2-1.7B and Qwen2.5-1.5B responses with 256 tokens at float16 and 4-bit
changes at most \GpLenChanged{} keyword labels per model and precision, all to refusal. In these
runs the 4-bit drop in refusal of SmolLM2-1.7B is \GpLenSmDropA{} points at both lengths,
against \SafeMaxDeltaPP{} in the baseline run on other hardware. Over-refusal is
measured with WildGuard alone. Its XSTest labels agree with the human label on \WgXsAnnAgree{} of
the \WgXsAnnN{} annotated responses it judged, too few to estimate its error on this dataset. Truthfulness and hallucination are measured as
preferences among given answers, so the study does not address hallucination in free
generation.

The benchmarks predate the models, and our contamination checks cover only AdvBench and only
verbatim exposure.
SmolLM2-135M, SmolLM2-360M and TinyLlama-1.1B refuse few harmful requests even at float16, so
their safety changes are bounded by a ceiling. SmolLM2-135M and SmolLM2-360M are also weak
general assistants, included to cover the lower end of the size range.

The protocol comparison uses four multiple-choice datasets and one lettered template, and other
templates and tasks may show different position effects. The batch-composition test uses 500
HaluEval items in \EoneBModels{} models, and the test of its cause \MechBatchN{} models for batch
arrangement and \MechPadN{} for padding, each with all of a model's settings on one machine: the laptop GPU for three models, and a cloud NVIDIA T4 for SmolLM2-135M, TinyLlama-1.1B and Gemma-3-1B, whose default 8-bit setting was rerun there. Gemma-3 is released in bfloat16 and is evaluated
in float16 for comparability. Scored in bfloat16 on the protocol items, its choices agree with
float16 on at least \GpBfAgreeMin\% of items per item set and rule. With bfloat16 as the
reference, the 4-bit change keeps its sign and significance in \GpBfSame{} of \GpBfCells{}
comparisons.

\section*{Ethics statement}

The harmful prompts come from public benchmarks, and the model outputs are released only for
research use. The two annotators were colleagues at the authors' university who took part
voluntarily, without payment. They were told of the harmful content in advance and saw only the
prompt and the response, not the model or the precision.

\bibliography{references}
\bibliographystyle{acl_natbib}

\appendix
\raggedbottom
\makeatletter
\setlength{\@fptop}{0pt}
\setlength{\@fpbot}{0pt plus 1fil}
\makeatother
\setcounter{table}{0}
\setcounter{figure}{0}
\renewcommand{\thetable}{A\arabic{table}}
\renewcommand{\thefigure}{A\arabic{figure}}

\section{Additional results}
\label{sec:extra}

\begin{figure}[!ht]
\centering
\includegraphics[width=\columnwidth]{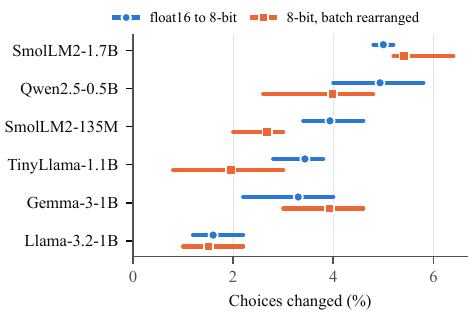}
\caption{Share of choices on 500 HaluEval items that change between float16 and LLM.int8()
(blue), and between two batch arrangements of the same 8-bit model (orange). Lines span the
arrangements compared; markers are means. Repeating an identical arrangement changes no choice,
and rearranging the batch at float16 changes at most \EoneBCrossSixteenHi\%.}
\label{fig:int8}
\end{figure}

Figure~\ref{fig:int8} gives the per-model batch test of Section~\ref{sec:int8}, and Table~\ref{tab:calibration} lists the calibration changes discussed in Section~\ref{sec:calibration}.

\begin{table*}[t]
\centering\small
\setlength{\tabcolsep}{8pt}
\begin{tabular}{l l c r r r}
\toprule
Model & Data & bit & $\Delta$acc & $\Delta$ECE {\scriptsize [95\% CI]} & recent. \\
\midrule
SmolLM2-135M & BoolQ & 4 & $-$10.7 & +0.154 {\scriptsize [+0.130, +0.173]} & $-$0.031 \\
Llama-3.2-1B & BoolQ & 4 & $-$6.1 & +0.037 {\scriptsize [+0.019, +0.051]} & $-$0.008 \\
Gemma-3-1B & HellaSwag & 4 & $-$1.8 & +0.023 {\scriptsize [+0.007, +0.035]} & -- \\
SmolLM2-360M & ARC-E & 4 & $-$1.9 & +0.019 {\scriptsize [+0.001, +0.037]} & -- \\
Qwen2.5-0.5B & PIQA & 8 & +0.9 & +0.009 {\scriptsize [+0.001, +0.016]} & +0.004 \\
TinyLlama-1.1B & PIQA & 8 & $-$0.8 & $-$0.007 {\scriptsize [$-$0.017, $-$0.001]} & $-$0.011 \\
TinyLlama-1.1B & PIQA & 4 & $-$1.6 & $-$0.012 {\scriptsize [$-$0.026, $-$0.001]} & $-$0.017 \\
SmolLM2-1.7B & BoolQ & 8 & +2.6 & $-$0.020 {\scriptsize [$-$0.032, $-$0.008]} & $-$0.001 \\
Gemma-3-1B & BoolQ & 4 & $-$5.7 & $-$0.020 {\scriptsize [$-$0.034, $-$0.006]} & $-$0.011 \\
TinyLlama-1.1B & BoolQ & 4 & +2.3 & $-$0.022 {\scriptsize [$-$0.031, $-$0.005]} & $-$0.002 \\
Qwen2.5-1.5B & BoolQ & 4 & $-$1.5 & $-$0.071 {\scriptsize [$-$0.083, $-$0.051]} & $-$0.055 \\
\bottomrule
\end{tabular}
\caption{The 11 of 80 calibration comparisons with float16 whose paired bootstrap interval
excludes zero. $\Delta$acc in points. Recent.: $\Delta$ECE after the cross-fitted intercept
recentring (binary tasks only).}
\label{tab:calibration}
\end{table*}

\section{Experimental details}
\label{sec:appendix}

\paragraph{Models and quantization.} The checkpoints, with the revision used in
Section~\ref{sec:protocol}, are HuggingFaceTB/SmolLM2-135M-Instruct (12fd25f),
SmolLM2-360M-Instruct (a10cc15), SmolLM2-1.7B-Instruct (31b70e2), Qwen/Qwen2.5-0.5B-Instruct
(7ae5576), Qwen2.5-1.5B-Instruct (989aa79), meta-llama/Llama-3.2-1B-Instruct (9213176),
google/gemma-3-1b-it (dcc83ea) and TinyLlama/TinyLlama-1.1B-Chat-v1.0 (fe8a4ea). The 8-bit
configuration is bitsandbytes \texttt{load\_in\_8bit} with the default outlier threshold of 6.0.
It quantizes weights and activations except for outlier dimensions. The 4-bit configuration is
NF4 without double quantization and with a float16 compute dtype. All unquantized weights are
float16. The second safety judge is allenai/wildguard (cbba482), run in float16 with greedy decoding of at most 32 tokens. For Gemma-3-1B, the AdvBench and XSTest responses were also generated in bfloat16, to check whether float16 degrades this model; these responses enter only the annotation sample.

\paragraph{Refusal detectors.} The baseline keyword detector lowercases the response and
reports a refusal if any of 30 markers occurs in its first 200 characters. The markers are ``i
cannot'', ``i can't'', ``i can not'', ``i won't'', ``i will not'', ``i'm sorry'', ``i am
sorry'', ``i apologize'', ``i'm not able'', ``i am not able'', ``i'm unable'', ``i am unable'',
``cannot help'', ``can't help'', ``can't assist'', ``cannot assist'', ``i must decline'', ``i'm
not comfortable'', ``i am not comfortable'', ``as an ai'', ``against my'', ``not appropriate'',
``i'm afraid i can'', ``unable to provide'', ``cannot provide'', ``can't provide'', ``will not
provide'', ``cannot fulfill'', ``can't fulfill'' and ``not able to help''. An empty response
counts as a refusal. The corrected detector uses the same markers. It maps typographic
apostrophes and quotation marks to their ASCII forms and scans the whole response.

\paragraph{Generation and calibration metrics.} Safety prompts are formatted with each model's
chat template as a single user turn. They are decoded greedily for at most 64 new tokens, in
left-padded batches of 8, with inputs truncated to 512 tokens. ECE uses 10 equal-width
confidence bins and ACE 10 equal-mass bins. Confidence is the probability of the chosen option
after a softmax over the options' mean log-probabilities.

\paragraph{Prompts.} For multiple-choice scoring, models receive plain text without a chat
template, as in the baseline evaluation. Each item has a stem: ``Question: \ldots'' for TruthfulQA and
ARC-Easy, ``Knowledge: \ldots\ Question: \ldots'' for HaluEval, and the context followed by
``Question: \ldots'' for BBQ. Under cloze scoring, each option is scored as the continuation of
the stem followed by a new line and ``Answer:''. The PMI variant subtracts the option's
log-probability after the prompt ``Answer:'' alone. Under lettered scoring, the prompt is the
stem, a new line, ``Options:'', one line per option of the form ``A. \textit{option}'', and
``Answer:''. The score of an option is the log-sum-exp of the next-token log-probabilities of
its letter with and without a leading space. Only forms that the model's vocabulary encodes as a
single token are included.

\paragraph{Items.} TruthfulQA uses all 817 validation questions. Both the single-answer and the
multiple-true-answer option sets are scored, so that standard MC1 and MC2 and the variants of
the baseline evaluation can be computed from the same run. HaluEval uses the 2,000 question-answering
items of the baseline evaluation, ARC-Easy 1,000 test questions sampled with a fixed seed, and BBQ the
first 1,000 items of the baseline evaluation's BBQ sample. Every option's summed log-probability, token
count and unconditional log-probability are stored, as is every lettered score in both orders.
All scoring rules are therefore computed offline from one pass per model and precision.

\paragraph{Batching and numerics.} Cloze and lettered prompts are grouped by length under a
fixed memory budget on the logits tensor. Splitting a float16 batch in two changed
log-probabilities by at most about 0.1 and no choice in a 360-item check. Arbitrary
rearrangement at float16 changes up to \EoneBCrossSixteenHi\% of choices
(Section~\ref{sec:int8}). The batch-composition test scores the (prompt, option) pairs of the
HaluEval items in four arrangements: original order in chunks of 16 pairs (8 items), sorted by
length, in a random order, and one pair at a time. The first and last arrangements are repeated.

\paragraph{Software and hardware.} We use PyTorch 2.8 with CUDA 12.8, Transformers 4.57 and
bitsandbytes 0.50. The protocol runs, the batch and padding tests and the regenerations ran on a single 6\,GB laptop GPU, and the baseline evaluation on an NVIDIA A10G; the mechanism test of Section~\ref{sec:int8} also used a cloud NVIDIA T4 for three models. The full protocol grid of 24 model and precision
cells and the batch-composition test take roughly a day of compute on this hardware.

\paragraph{Contamination checks.}
\label{sec:contamination}
Both checks run each model at float16 on the 520 AdvBench instructions. The exchangeability
test \citep{oren2024proving} splits the instructions into 10 shards and compares, within each
shard, the log-likelihood of the released order with that of 30 random orders. A one-sided
$t$-test on the standardised gaps gives one $p$ per model. Min-K\% Prob \citep{shi2024detecting}
averages the lowest 20\% of token log-probabilities of an instruction. Its AUROC is computed for
separating refused from other items and items whose outcome flips at 4-bit from the rest, with a
Mann--Whitney test per model. Holm correction is applied over the \NModels{} models within each
test. Table~\ref{tab:contamination} gives the results.

\begin{table}[t]
\centering\small
\resizebox{\columnwidth}{!}{%
\begin{tabular}{l rr rr rr}
\toprule
& \multicolumn{2}{c}{Exchangeability} & \multicolumn{2}{c}{float16 refusal} & \multicolumn{2}{c}{4-bit flip} \\
\cmidrule(lr){2-3}\cmidrule(lr){4-5}\cmidrule(lr){6-7}
Model & $\bar z$ & $p$ & \% & AUROC & \% & AUROC \\
\midrule
SmolLM2-135M & $-$0.03 & 0.54 & 4.8 & 0.57 & 4.4 & 0.51 \\
SmolLM2-360M & $-$0.48 & 0.84 & 14.4 & 0.54 & 11.0 & 0.50 \\
Qwen2.5-0.5B & +0.16 & 0.45 & 91.5 & 0.49 & 5.2 & 0.47 \\
Gemma-3-1B & $-$0.21 & 0.87 & 90.6 & 0.52 & 17.7 & 0.47 \\
TinyLlama-1.1B & $-$0.23 & 0.81 & 1.2 & 0.29 & 2.7 & 0.32 \\
Llama-3.2-1B & +0.16 & 0.29 & 98.3 & 0.34 & 0.8 & 0.51 \\
Qwen2.5-1.5B & $-$0.07 & 0.54 & 99.8 & 0.59 & 0.8 & 0.47 \\
SmolLM2-1.7B & $-$0.45 & 1.00 & 63.1 & 0.52 & 28.1 & 0.47 \\
\bottomrule
\end{tabular}}
\caption{Contamination checks on AdvBench at float16. Exchangeability: mean standardised gap
between the log-likelihood of the instructions in their released order and in 30 random orders
over 10 shards, with a one-sided $t$-test \citep{oren2024proving}. AUROC: how well an item's
Min-K\% Prob \citep{shi2024detecting} separates refused from other items, and items whose
outcome flips at 4-bit from the rest (0.5 = no relation). No test is significant after Holm
correction.}
\label{tab:contamination}
\end{table}

\paragraph{Reproduction of the baseline evaluation.} Table~\ref{tab:repro} gives, per model and
precision, the share of items on which the protocol runs pick the same cloze option as the
baseline evaluation. The two runs use different hardware and different batches.

\begin{table}[!t]
\centering\small
\setlength{\tabcolsep}{3.5pt}
\resizebox{\columnwidth}{!}{%
\begin{tabular}{l rr rr rr}
\toprule
& \multicolumn{2}{c}{float16} & \multicolumn{2}{c}{8-bit} & \multicolumn{2}{c}{4-bit} \\
\cmidrule(lr){2-3}\cmidrule(lr){4-5}\cmidrule(lr){6-7}
Model & Halu & BBQ & Halu & BBQ & Halu & BBQ \\
\midrule
SmolLM2-135M & 100.0 & 99.9 & 96.2 & 92.0 & 100.0 & 99.8 \\
SmolLM2-360M & 99.9 & 99.9 & 96.5 & 89.6 & 100.0 & 99.7 \\
Qwen2.5-0.5B & 99.8 & 99.8 & 96.8 & 92.4 & 99.9 & 99.6 \\
Gemma-3-1B & 99.9 & 100.0 & 96.2 & 91.4 & 100.0 & 99.7 \\
TinyLlama-1.1B & 100.0 & 99.8 & 98.2 & 96.0 & 100.0 & 99.9 \\
Llama-3.2-1B & 100.0 & 99.8 & 98.2 & 94.4 & 99.9 & 99.5 \\
Qwen2.5-1.5B & 100.0 & 99.8 & 97.0 & 94.2 & 100.0 & 99.7 \\
SmolLM2-1.7B & 99.9 & 100.0 & 92.8 & 90.9 & 99.9 & 100.0 \\
\bottomrule
\end{tabular}}
\caption{Share (\%) of items on which the protocol runs of Section~\ref{sec:protocol} pick the same
cloze option as the baseline evaluation (2,000 HaluEval and 1,000 BBQ items, matched by item hash).}
\label{tab:repro}
\end{table}

\end{document}